\documentclass[12]{article}
\usepackage{hyperref}
\usepackage{amsmath}
\usepackage{graphicx}
\usepackage{amsfonts}
\usepackage{amssymb}
\usepackage{algorithm}
\usepackage{algorithmic}

\begin{document}

\title{A Siamese Deep Forest}
\author{Lev V. Utkin$^{1}$ and Mikhail A. Ryabinin$^{2}$\\Department of Telematics\\Peter the Great St.Petersburg Polytechnic University\\St.Petersburg, Russia\\e-mail: $^{1}$lev.utkin@gmail.com, $^{2}$mihail-ryabinin@yandex.ru}
\date{}
\maketitle

\begin{abstract}
A Siamese Deep Forest (SDF) is proposed in the paper. It is based on the Deep
Forest or gcForest proposed by Zhou and Feng and can be viewed as a gcForest
modification. It can be also regarded as an alternative to the well-known
Siamese neural networks. The SDF uses a modified training set consisting of
concatenated pairs of vectors. Moreover, it defines the class distributions in
the deep forest as the weighted sum of the tree class probabilities such that
the weights are determined in order to reduce distances between similar pairs
and to increase them between dissimilar points. We show that the weights can
be obtained by solving a quadratic optimization problem. The SDF aims to
prevent overfitting which takes place in neural networks when only limited
training data are available. The numerical experiments illustrate the proposed
distance metric method.


\end{abstract}

\section{Introduction}

One of the important machine learning tasks is to compare pairs of objects,
for example, pairs of images, pairs of data vectors, etc. There are a lot of
approaches for solving the task. One of the approaches is based on computing a
corresponding pairwise metric function which measures a distance between data
vectors or a similarity between the vectors. This approach is called the
metric learning \cite{Bellet-etal-2013,Kulis-2012,Zheng-etal-2016}. It is
pointed out by Bellet et al. \cite{Bellet-etal-2013} in their review paper
that the metric learning aims to adapt the pairwise real-valued metric
function, for example, the Mahalanobis distance or the Euclidean distance, to
a problem of interest using the information provided by training data. A
detailed description of the metric learning approaches is also represented by
Le Capitaine \cite{LeCapitaine-2016} and by Kulis \cite{Kulis-2012}. The basic
idea underlying the metric learning solution is that the distance between
similar objects should be smaller than the distance between different objects.

Suppose there is a training set $S$ $=\{(\mathbf{x}_{i},\mathbf{x}_{j}%
,y_{ij}),\ (i,j)\in K\}$ consisting of $N$ pairs of examples $\mathbf{x}%
_{i}\in \mathbb{R}^{m}$ and $\mathbf{x}_{j}\in \mathbb{R}^{m}$ such that a
binary label $y_{ij}\in \{0,1\}$ is assigned to every pair $(\mathbf{x}%
_{i},\mathbf{x}_{j})$. If two data vectors $\mathbf{x}_{i}$ and $\mathbf{x}%
_{j}$ are semantically similar or belong to the same class of objects, then
$y_{ij}$ takes the value $0$. If the vectors correspond to different or
semantically dissimilar objects, then $y_{ij}$ takes the value $1$. This
implies that the training set $S$ can be divided into two subsets. The first
subset is called the similar or positive set and is defined as
\[
\mathcal{S}=\{(\mathbf{x}_{i},\mathbf{x}_{j}):\mathbf{x}_{i}\text{ and
}\mathbf{x}_{j}\  \text{are semantically similar and }y_{ij}=0\}.
\]

The second subset is the dissimilar or negative set. It is defined as
\[
\mathcal{D}=\{(\mathbf{x}_{i},\mathbf{x}_{j}):\mathbf{x}_{i}\text{ and
}\mathbf{x}_{j}\  \text{are semantically dissimilar and }y_{ij}=1\}.
\]

If we have two observation vectors $\mathbf{x}_{i}\in \mathbb{R}^{m}$ and
$\mathbf{x}_{j}\in \mathbb{R}^{m}$ from the training set, then the distance
$d(\mathbf{x}_{i},\mathbf{x}_{j})$ should be minimized if $\mathbf{x}_{i}$ and
$\mathbf{x}_{j}$ are semantically similar, and it should be maximized between
dissimilar $\mathbf{x}_{i}$ and $\mathbf{x}_{j}$. The most general and popular
real-valued metric function is the squared Mahalanobis distance $d_{M}%
^{2}(\mathbf{x}_{i},\mathbf{x}_{j})$ which is defined for vectors
$\mathbf{x}_{i}$ and $\mathbf{x}_{j}$ as
\[
d_{M}^{2}(\mathbf{x}_{i},\mathbf{x}_{j})=(\mathbf{x}_{i}-\mathbf{x}%
_{j})^{\mathrm{T}}M(\mathbf{x}_{i}-\mathbf{x}_{j}).
\]

Here $M\in \mathbb{R}^{m\times m}$ is a symmetric positive semi-defined matrix.
If $\mathbf{x}_{i}$ and $\mathbf{x}_{j}$ are random vectors from the same
distribution with covariance matrix $C$, then $M=C^{-1}$. If $M$ is the
identity matrix, then $d_{M}^{2}(\mathbf{x}_{i},\mathbf{x}_{j})$ is the
squared Euclidean distance.

Given subsets $\mathcal{S}$ and $\mathcal{D}$, the metric learning
optimization problem can be formulated as follows:
\[
M^{\ast}=\arg \min_{M}\left[  J(M,\mathcal{D},\mathcal{S})+\lambda \cdot
R(M)\right]  ,
\]
where $J(M,\mathcal{D},\mathcal{S})$ is a loss function that penalizes
violated constraints; $R(M)$ is some regularizer on $M$; $\lambda \geq0$ is the
regularization parameter.

There are many useful loss functions $J$ which take into account the condition
that the distance between similar objects should be smaller than the distance
between different objects. These functions define a number of learning
methods. It should be noted that the learning methods using the Mahalanobis
distance assume some linear structure of data. If this is not valid, then the
kernelization of linear methods is one of the possible ways for solving the
metric learning problem. Bellet et al. \cite{Bellet-etal-2013} review several
approaches and algorithms to deal with nonlinear forms of metrics. In
particular, these are the Support Vector Metric Learning algorithm provided by
Xu et al. \cite{Xu-Weinberger-Chapelle-2012}, the Gradient-Boosted Large
Margin Nearest Neighbors method proposed by Kedem et al.
\cite{Kedem-etal-2012}, the Hamming Distance Metric Learning algorithm
provided by Norouzi et al. \cite{Norouzi-etal-2012}.

A powerful implementation of the metric learning dealing with non-linear data
structures is the so-called Siamese neural network introduced by Bromley et
al. \cite{Bromley-etal-1993} in order to solve signature verification as a
problem of image matching. This network consists of two identical sub-networks
joined at their outputs. The two sub-networks extract features from two input
examples during training, while the joining neuron measures the distance
between the two feature vectors. The Siamese architecture has been exploited
in many applications, for example, in face verification
\cite{Chopra-etal-2005}, in the one-shot learning in which predictions are
made given only a single example of each new class \cite{Koch-etal-2015}, in
constructing an inertial gesture classification \cite{Berlemont-etal-2015}, in
deep learning \cite{Wang-etal-2016}, in extracting speaker-specific
information \cite{Chen-Salman-2011}, for face verification in the wild
\cite{Hu-Lu-Tan-2014}. This is only a part of successful applications of
Siamese neural networks. Many modifications of Siamese networks have been
developed, including fully-convolutional Siamese networks
\cite{Bertinetto-etal-2016}, Siamese networks combined with a gradient
boosting classifier \cite{Leal-Taixe-etal-2016}, Siamese networks with the
triangular similarity metric \cite{Zheng-etal-2016}.

One of the difficulties of the Siamese neural network as well as other neural
networks is that limited training data lead to overfitting when training
neural networks. Many different methods have been developed to prevent
overfitting, for example, dropout methods \cite{Srivastava-etal-2014} which
are based on combination of the results of different networks by randomly
dropping out neurons in the network. A very interesting new method which can
be regarded as an alternative to deep neural networks is the deep forest
proposed by Zhou and Feng \cite{Zhou-Feng-2017} and called the gcForest. In
fact, this is a multi-layer structure where each layer contains many random
forests, i.e., this is an ensemble of decision tree ensembles. Zhou and Feng
\cite{Zhou-Feng-2017} point out that their approach is highly competitive to
deep neural networks. In contrast to deep neural networks which require great
effort in hyperparameter tuning and large-scale training data, gcForest is
much easier to train and can perfectly work when there are only small-scale
training data. The deep forest solves tasks of classification as well as
regression. Therefore, by taking into account its advantages, it is important
to modify it in order to develop a structure solving the metric learning task.
We propose the so-called Siamese Deep Forest (SDF) which can be regarded as an
alternative to the Siamese neural networks and which is based on the gcForest
proposed by Zhou and Feng \cite{Zhou-Feng-2017} and can be viewed as its
modification. Three main ideas underlying the SDF can be formulated as follows:

\begin{enumerate}
\item We propose to modify training set by using concatenated pairs of vectors.

\item We define the class distributions in the deep forest as the weighted sum
of the tree class probabilities where the weights are determined in order to
reduce distances between semantically similar pairs of examples and to
increase them between dissimilar pairs. The weights are training parameters of
the SDF.

\item We apply the greedy algorithm for training the SDF, i.e., the weights
are successively computed for every layer or level of the forest cascade.
\end{enumerate}

We consider the case of the weakly supervised learning \cite{Bellet-etal-2013}
when there are no information about the class labels of individual training
examples, but only information in the form of sets $\mathcal{S}$ and
$\mathcal{D}$ is provided, i.e., we know only semantic similarity of pairs of
training data. However, the case of the fully supervised learning when the
class labels of individual training examples are known can be considered in
the same way.

It should be noted that the SDF cannot be called Siamese in the true sense of
the word. It does not consist of two gcForests like the Siamese neural
network. However, its aim coincides with the Siamese network aim. Therefore,
we give this name for the gcForest modification.

The paper is organized as follows. Section 2 gives a very short introduction
into the Siamese neural networks. A short description of the gcForest proposed
by Zhou and Feng \cite{Zhou-Feng-2017} is given in Section 3. The ideas
underlying the SDF are represented in Section 4 in detail. A modification of
the gcForest using the weighted averages, which can be regarded as a basis of
the SDF is provided in Section 5. Algorithms for training and testing the SDF
are considered in Section 6. Numerical experiments with real data illustrating
cases when the proposed SDF outperforms the gcForest are given in Section 7.
Concluding remarks are provided in Section 8.

\section{Siamese neural networks}

Before studying the SDF, we consider the Siamese neural network which is an
efficient and popular tool for dealing with data of the form $\mathcal{S}$ and
$\mathcal{D}$. It will be a basis for constructing the SDF.

\begin{figure}
[ptb]
\begin{center}
\includegraphics[
%natheight=2.006600in,
%natwidth=2.675200in,
height=1.5805in,
width=2.0997in
]%
{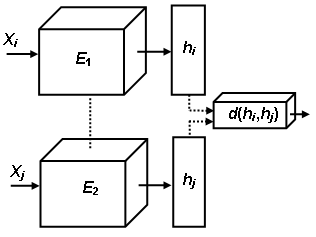}%
\caption{An architecture of the Siamese neural network}%
\label{fig:siamese_net}%
\end{center}
\end{figure}

A standard architecture of the Siamese network given in the literature (see,
for example, \cite{Chopra-etal-2005}) is shown in Fig. \ref{fig:siamese_net}.
Let $\mathbf{x}_{i}$ and $\mathbf{x}_{j}$ be two data vectors corresponding to
a pair of elements from a training set, for example, images. Suppose that $f$
is a map of $\mathbf{x}_{i}$ and $\mathbf{x}_{j}$ to a low-dimensional space
such that it is implemented as a neural network with the weight matrix $W$. At
that, parameters $W$ are shared by two neural networks $f(\mathbf{x}_{1})$ and
$f(\mathbf{x}_{2})$ denoted as $E_{1}$ and $E_{2}$ and corresponding to
different input vectors, i.e., they are the same for the two neural networks.
The property of the same parameters in the Siamese neural network is very
important because it defines the corresponding training algorithm. By
comparing the outputs $\mathbf{h}_{i}=f(\mathbf{x}_{i})$ and $\mathbf{h}%
_{j}=f(\mathbf{x}_{j})$ using the Euclidean distance $d(\mathbf{h}%
_{i},\mathbf{h}_{j})$, we measure the compatibility between $\mathbf{x}_{i}$
and $\mathbf{x}_{j}$.
%TCIMACRO{\FRAME{ftbpFU}{2.0997in}{1.5805in}{0pt}{\Qcb{An architecture of the
%Siamese neural network}}{\Qlb{fig:siamese_net}}{siamese_network_1.png}%
%width 2.0997in;  height 1.5805in;  depth 0pt;  original-width 2.6752in;
%original-height 2.0066in;  cropleft "0";  croptop "1";  cropright "1";
%filename 'Siamese_network_1.png';file-properties "XNPEU";}%
%}}%
%BeginExpansion
%EndExpansion

If we assume for simplicity that the neural network has one hidden layer, then
there holds
\[
\mathbf{h}=\sigma(W\mathbf{x}+b).
\]
Here $\sigma(z)$ is an activation function; $W$ is the weight $p\times M$
matrix such that its element $w_{ij}$ is the weight of the connection between
unit $j$ in the input layer and unit $i$ in the hidden layer, $i=1,...,p$,
$j=1,...,M$; $b=(b_{1},...,b_{p})$ is a bias vector; $\mathbf{h}%
=(h_{1},...,h_{p})$ is the vector of neuron activations, which depends on the
input vector $\mathbf{x}$.

The Siamese neural network is trained on pairs of observations by using
specific loss functions, for example, the following contrastive loss
function:
\begin{equation}
l(\mathbf{x}_{i},\mathbf{x}_{j},y_{ij})=\left \{
\begin{array}
[c]{cc}%
\left \Vert \mathbf{h}_{i}-\mathbf{h}_{j}\right \Vert _{2}^{2}, & y_{ij}=0,\\
\max(0,\tau-\left \Vert \mathbf{h}_{i}-\mathbf{h}_{j}\right \Vert _{2}^{2}), &
y_{ij}=1,
\end{array}
\right.  \label{SiamDF_20}%
\end{equation}
where $\tau$ is a predefined threshold.

Hence, the total error function for minimizing is defined as
\[
J(W,b)=\sum \nolimits_{i,j}l(\mathbf{x}_{i},\mathbf{x}_{j},y_{ij})+\mu R(W,b).
\]

Here $R(W,b)$ is a regularization term added to improve generalization of the
neural network, $\mu$ is a hyper-parameter which controls the strength of the
regularization. The above problem can be solved by using the stochastic
gradient descent scheme.

\section{Deep Forest}

According to \cite{Zhou-Feng-2017}, the gcForest generates a deep forest
ensemble, with a cascade structure. Representation learning in deep neural
networks mostly relies on the layer-by-layer processing of raw features. The
gcForest representational learning ability can be further enhanced by the
so-called multi-grained scanning. Each level of cascade structure receives
feature information processed by its preceding level, and outputs its
processing result to the next level. Moreover, each cascade level is an
ensemble of decision tree forests. We do not consider in detail the
Multi-Grained Scanning where sliding windows are used to scan the raw features
because this part of the deep forest is the same in the SDF. However, the most
interesting component of the gcForest from the SDF construction point of view
is the cascade forest.%

%TCIMACRO{\FRAME{ftbpFU}{4.1537in}{1.8502in}{0pt}{\Qcb{The architecture of the
%cascade forest \cite{Zhou-Feng-2017}}}{\Qlb{fig:cascade_forest}}%
%{forest_cascade.png}{\special{ language "Scientific Word";  type "GRAPHIC";
%width 4.1537in;  height 1.8502in;  depth 0pt;  original-width 9.3292in;
%original-height 4.1397in;  cropleft "0";  croptop "1";  cropright "1";
%cropbottom "0";  filename 'forest_cascade.png';file-properties "XNPEU";}}}%
%BeginExpansion
\begin{figure}
[ptb]
\begin{center}
\includegraphics[
%natheight=4.139700in,
%natwidth=9.329200in,
height=1.8502in,
width=4.1537in
]%
{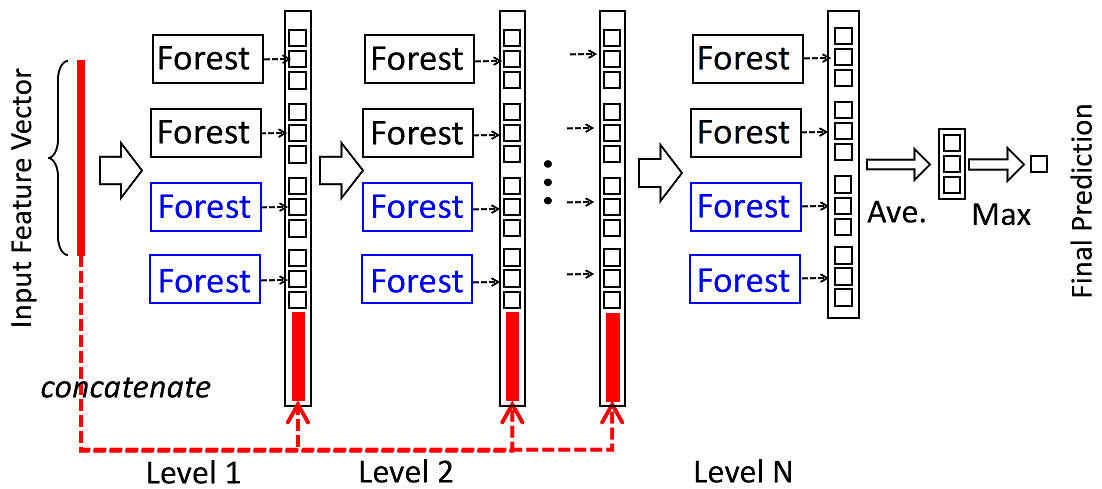}%
\caption{The architecture of the cascade forest \protect\cite{Zhou-Feng-2017}}%
\label{fig:cascade_forest}%
\end{center}
\end{figure}
%EndExpansion

Given an instance, each forest produces an estimate of class distribution by
counting the percentage of different classes of examples at the leaf node
where the concerned instance falls into, and then averaging across all trees
in the same forest. The class distribution forms a class vector, which is then
concatenated with the original vector to be input to the next level of
cascade. The usage of the class vector as a result of the random forest
classification is very similar to the idea underlying the stacking method
\cite{Wolpert-1992}. The stacking algorithm trains the first-level learners
using the original training data set. Then it generates a new data set for
training the second-level learner (meta-learner) such that the outputs of the
first-level learners are regarded as input features for the second-level
learner while the original labels are still regarded as labels of the new
training data. In fact, the class vectors in the gcForest can be viewed as the
meta-learners. In contrast to the stacking algorithm, the gcForest
simultaneously uses the original vector and the class vectors (meta-learners)
at the next level of cascade by means of their concatenation. This implies
that the feature vector is enlarged and enlarged after every cascade level.
The architecture of the cascade proposed by Zhou and Feng
\cite{Zhou-Feng-2017} is shown in Fig. \ref{fig:cascade_forest}. It can be
seen from the figure that each level of the cascade consists of two different
pairs of random forests which generate 3-dimensional class vectors
concatenated each other and with the original input. After the last level, we
have the feature representation of the input feature vector, which can be
classified in order to get the final prediction. Zhou and Feng
\cite{Zhou-Feng-2017} propose to use different forests at every level in order
to provide the diversity which is an important requirement for the random
forest construction.

\section{Three ideas underlying the SDF}

The SDF aims to function like the standard Siamese neural network. This
implies that the SDF should provide large distances between semantically
similar pairs of vectors and small distances between dissimilar pairs. We
propose three main ideas underlying the SDF:

\begin{enumerate}
\item Denote the set indices of all pairs $\mathbf{x}_{i}$ and $\mathbf{x}%
_{j}$ as $K=\{(i,j)\}$ We train every tree by using the concatenation of two
vectors $\mathbf{x}_{i}$ and $\mathbf{x}_{j}$ such that the class $y_{ij}%
\in \{0,1\}$ is defined by the semantical similarity of the vectors. In fact,
the trees are trained on the basis of two classes and reflect the semantical
similarity of pairs, but not classes of separate examples. With this
concatenation, we define a new set of classes such that we do not need to know
separate classes for $\mathbf{x}_{i}$ or for $\mathbf{x}_{j}$. As a result, we
have a new training set $R=\{(\mathbf{x}_{i},\mathbf{x}_{j}),y_{ij}%
),\ (i,j)\in K\}$ and exploit only the information about the semantical
similarity. The concatenation is not necessary when the classes of training
elements are known, i.e., we have a set of labels $\{y_{1},...,y_{n}\}$. In
this case, only the second idea can be applied.

\item We partially use some modification of ideas provided by Xiong et al.
\cite{Xiong-etal-2012} and Dong et al. \cite{Dong-Du-Zhang-2015}. In
particular, Xiong et al. \cite{Xiong-etal-2012} considered an algorithm for
solving the metric learning problem by means of the random forests. The
proposed metric is able to implicitly adapt its distance function throughout
the feature space. Dong et al. \cite{Dong-Du-Zhang-2015} proposed a random
forest metric learning (RFML) algorithm, which combines semi-multiple metrics
with random forests to better separate the desired targets and background in
detecting and identifying target pixels based on specific spectral signatures
in hyperspectral image processing. A common idea underlying the metric
learning algorithms in \cite{Xiong-etal-2012} and \cite{Dong-Du-Zhang-2015} is
that the distance measure between a pair of training elements $\mathbf{x}%
_{i},\mathbf{x}_{j}$ for a combination of trees is defined as average of some
special functions of the training elements. For example, if a random forest is
a combination of $T$ decision trees $\{f_{t}(\mathbf{x}),t=1,...,T\}$, then
the distance measure is
\[
d(\mathbf{x}_{i},\mathbf{x}_{j})=T^{-1}\sum_{t=1}^{T}f_{t}(\psi(\mathbf{x}%
_{i},\mathbf{x}_{j})).
\]
Here $\psi(\mathbf{x}_{i},\mathbf{x}_{j})$ is a mapping function which is
specifically defined in \cite{Xiong-etal-2012} and \cite{Dong-Du-Zhang-2015}.
We combine the above ideas with the idea of probability distributions of
classes provided in \cite{Zhou-Feng-2017} in order to produce a new feature
vector after every level of the cascade forest. According to
\cite{Zhou-Feng-2017}, each forest of a cascade level produces an estimate of
the class probability distribution by counting the percentage of different
classes of training examples at the leaf node where the concerned instance
falls into, and then averaging across all trees in the same forest. Our idea
is to define the forest class distribution as a weighted sum of the tree class
probabilities. At that, the weights are computed in an optimal way in order to
reduce distances between similar pairs and to increase them between dissimilar points.

The obtained weights are very similar to weights of the neural network
connections between neurons, which are also computed during training the
neural network. The trained values of weights in the SDF are determined in
accordance with a loss function defining properties of the SDF or the neural
network. Due to this similarity, we will call levels of the cascade as layers sometimes.

It should be also noted that the first idea can be sufficient for implementing
the SDF because the additional features (the class vectors) produced by the
previous cascade levels partly reflect the semantical similarity of pairs of
examples. However, in order to enhance the discriminative capability of the
SDF, we modify the corresponding class distributions.

\item We apply the greedy algorithm for training the SDF that is we train
separately every level starting from the first level such that every next
level uses results of training at the previous level. In contrast to many
neural networks, the weights considered above are successively computed for
every layer or level of the forest cascade.
\end{enumerate}

\section{The SDF construction}

Let us introduce notations for indices corresponding to different deep forest
components. The indices and their sets of values are shown in Table
\ref{t:SiamDF_1}. One can see from Table \ref{t:SiamDF_1}, that there are $Q$
levels of the deep forest or the cascade, every level contains $M_{q}$ forests
such that every forest consists of $T_{k,q}$ trees. If we use the
concatenation of two vectors $\mathbf{x}_{i}$ and $\mathbf{x}_{j}$ for
defining new classes of semantically similar and dissimilar pairs, then the
number of classes is $2$. It should be noted that the class $c$ corresponds to
label $y_{ij}\in \{0,1\}$ of a training example from the set $R$.%

%TCIMACRO{\TeXButton{B}{\begin{table}[tbp] \centering}}%
%BeginExpansion
\begin{table}[tbp] \centering
%EndExpansion
\caption{Notations for indices}%
\begin{tabular}
[c]{cc}\hline
type & index\\ \hline
cascade level & $q=1,...,Q$\\ \hline
forest & $k=1,...,M_{q}$\\ \hline
tree & $t=1,...,T_{k,q}$\\ \hline
class & $c=0,1$\\ \hline
\end{tabular}
\label{t:SiamDF_1}%
%TCIMACRO{\TeXButton{E}{\end{table}}}%
%BeginExpansion
\end{table}%
%EndExpansion

Suppose we have trained trees in the SDF. One of the approaches underlying the
deep forest is that the class distribution forms a class vector which is then
concatenated with the original vector to be an input to the next level of the
cascade. Suppose a pair of the original vectors is $(\mathbf{x}_{i}%
,\mathbf{x}_{j})$, and the $p_{ij,c}^{(t,k,q)}$ is the probability of class
$c$ for the pair $(\mathbf{x}_{i},\mathbf{x}_{j})$ produced by the $t$-th tree
from the $k$-th forest at the cascade level $q$. Below we use the triple index
$(t,k,q)$ in order to indicate that the element belongs to the $t$-th tree
from the $k$-th forest at the cascade level $q$. The same can be said about
subsets of the triple. Then, according to \cite{Zhou-Feng-2017}, the element
$v_{c}^{(k,q)}$ of the class vector corresponding to class $c$ and produced by
the $k$-th forest in the gcForest is determined as%
\[
v_{ij,c}^{(k,q)}=T_{k,q}^{-1}\sum_{t=1}^{T_{k,q}}p_{ij,c}^{(t,k,q)}.
\]

Denote the obtained class vector as $\mathbf{v}_{ij}^{(k,q)}=(v_{ij,0}%
^{(k,q)},v_{ij,1}^{(k,q)})$. Then the concatenated vector $\mathbf{x}%
_{ij}^{(1)}$ after the first level of the cascade is
\[
\mathbf{x}_{ij}^{(1)}=\left(  \mathbf{x}_{i},\mathbf{x}_{j},\mathbf{v}%
_{ij}^{(1,1)},....,\mathbf{v}_{ij}^{(M_{1},1)}\right)  =\left(  \mathbf{x}%
_{i},\mathbf{x}_{j},\mathbf{v}_{ij}^{(k,1)},k=1,...,M_{1}\right)  .
\]
It is composed of the original vectors $\mathbf{x}_{i}$, $\mathbf{x}_{j}$ and
$M_{1}$ class vectors obtained from $M_{1}$ forests at the first level. In the
same way, we can write the concatenated vector $\mathbf{x}_{ij}^{(q)}$ after
the $q$-th level of the cascade as
\begin{align}
\mathbf{x}_{ij}^{(q)}  &  =\left(  \mathbf{x}_{i}^{(q-1)},\mathbf{x}%
_{j}^{(q-1)},\mathbf{v}_{ij}^{(1,q)},....,\mathbf{v}_{ij}^{(M_{q},q)}\right)
\nonumber \\
&  =\left(  \mathbf{x}_{i}^{(q-1)},\mathbf{x}_{j}^{(q-1)},\mathbf{v}%
_{ij}^{(k,q)},\ k=1,...,M_{q}\right)  . \label{SiamDF_40}%
\end{align}

In order to reduce the number of indices, we omit the index $q$ below because
all derivations will concern only level $q$, where $q$ may be arbitrary from
$1$ to $Q$. We also replace notations $M_{q}$ and $T_{k,q}$ with $M$ and
$T_{k}$, respectively, assuming that the number of forests and numbers of
trees strongly depend on the cascade level.

\begin{figure}
[ptb]
\begin{center}
\includegraphics[
%natheight=2.277200in,
%natwidth=9.126300in,
height=1.4355in,
width=5.7in
]%
{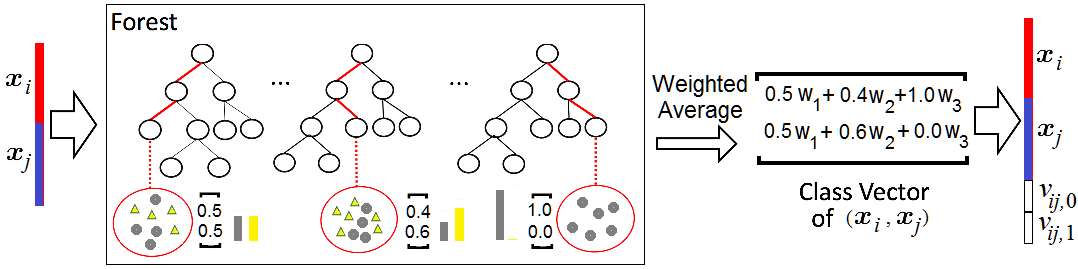}%
\caption{An illustration of the class vector generation taking into account
the weights}%
\label{fig:weighted_class}%
\end{center}
\end{figure}

The vector $\mathbf{x}_{ij}$ in (\ref{SiamDF_40}) has been derived in
accordance with the gcForest algorithm \cite{Zhou-Feng-2017}. However, in
order to implement the SDF, we propose to change the method for computing
elements $v_{ij,c}^{(k)}$ of the class vector, namely, the averaging is
replaced with the weighted sum of the form:
\begin{equation}
v_{ij,c}^{(k)}=\sum_{t=1}^{T_{k}}p_{ij,c}^{(t,k)}w^{(t,k)}. \label{SiamDF_41}%
\end{equation}
Here $w^{(t,k)}$ is a weight for combining the class probabilities of the
$t$-th tree from the $k$-th forest at the cascade level $q$. The weights play
a key role in implementing the SDF. An illustration of the weighted averaging
is shown in Fig. \ref{fig:weighted_class}, where we partly modify a picture
from \cite{Zhou-Feng-2017} (the left part is copied from \cite[Fig.
2]{Zhou-Feng-2017}) in order to show how elements of the class vector are
derived as a simple weighted sum. It can be seen from Fig.
\ref{fig:weighted_class} that two-class distribution is estimated by counting
the percentage of different classes ($y_{ij}=0$ or $y_{ij}=1$) of new training
concatenated examples $\left(  \mathbf{x}_{i},\mathbf{x}_{j}\right)  $ at the
leaf node where the concerned example $\left(  \mathbf{x}_{i},\mathbf{x}%
_{j}\right)  $ falls into. Then the class vector of $\left(  \mathbf{x}%
_{i},\mathbf{x}_{j}\right)  $ is computed as the weighted average. It is
important to note that we weigh trees belonging to one of the forests, but not
classes, i.e., the weights do not depend on the class $c$. Moreover, the
weights characterize trees, but not training elements. This implies that they
do not depend on the vectors $\mathbf{x}_{i}$, $\mathbf{x}_{j}$ too. One can
also see from Fig. \ref{fig:weighted_class} that the augmented features
$v_{ij,0}^{(k)}$ and $v_{ij,1}^{(k)}$ or the class vector corresponding to the
$k$-th forest are obtained as weighted sums, i.e., there hold
\begin{align*}
v_{ij,0}^{(k)}  &  =0.5\cdot w^{(1,k)}+0.4\cdot w^{(2,k)}+1\cdot w^{(3,k)},\\
v_{ij,1}^{(k)}  &  =0.5\cdot w^{(1,k)}+0.6\cdot w^{(2,k)}+0\cdot w^{(3,k)}.
\end{align*}

The weights are restricted by the following obvious condition:
\begin{equation}
\sum_{t=1}^{T_{k}}w^{(t,k)}=1. \label{SiamDF_42}%
\end{equation}

In other words, we have the weighted averages for every forest, and the
corresponding weights can be regarded as trained parameters in order to
decrease the distance between semantically similar $\mathbf{x}_{i}$ and
$\mathbf{x}_{j}$ and to increase the distance between dissimilar
$\mathbf{x}_{i}$ and $\mathbf{x}_{j}$. Therefore, we have to develop a way for
training the SDF, i.e., for computing the weights for every forest and for
every cascade level.%

%TCIMACRO{\FRAME{ftbpFU}{5.7in}{1.4355in}{0pt}{\Qcb{An illustration of the
%class vector generation taking into account the weights}}%
%{\Qlb{fig:weighted_class}}{weighted_class_vector_gen_2.png}%
%maintain-aspect-ratio TRUE;  display "USEDEF";  valid_file "F";  width 5.7in;
%height 1.4355in;  depth 0pt;  original-width 9.1263in;
%original-height 2.2772in;  cropleft "0";  croptop "1";  cropright "1";
%filename 'Weighted_Class_vector_gen_2.png';file-properties "XNPEU";}}}%
%BeginExpansion
%EndExpansion

Now we have numbers $v_{ij,c}^{(k)}$ for every class. Let us analyze these
numbers from the point of the SDF aim view.

First, we consider the case when $(\mathbf{x}_{i},\mathbf{x}_{j}%
)\in \mathcal{S}$ and $y_{ij}=0$. However, we may have non-zero $v_{ij,c}%
^{(k)}$ for both classes. It is obvious that $v_{ij,0}^{(k)}$ (the average
probability of class $c=0$) should be as large as possible because
$c=y_{ij}=0$. Moreover, $v_{ij,1}^{(k)}$ (the average probability of class
$c=1$) should be as small as possible because $c\neq y_{ij}=0$.

We can similarly write conditions for the case when $(\mathbf{x}%
_{i},\mathbf{x}_{j})\in \mathcal{D}$ and $y_{ij}=1$. In this case,
$v_{ij,0}^{(k)}$ should be as small as possible because $c\neq y_{ij}=1$, and
$v_{ij,1}^{(k)}$ should be as large as possible because $c=y_{ij}=1$.

In sum, we should increase (decrease) $v_{ij,c}^{(k)}$ if $c=y_{ij}$ ($c\neq
y_{ij}$). In other words, we have to find the weights maximizing (minimizing)
$v_{ij,c}^{(k)}$ when $c=y_{ij}$ ($c\neq y_{ij}$). The ideal case is when
$v_{ij,c}^{(k)}=1$ by $c=y_{ij}$ and $v_{ij,c}^{(k)}=0$ by $c\neq y_{ij}$.
However, the vector of weights has to be the same for every class, and it does
not depend on a certain class. At first glance, we could find optimal weights
for every individual forest separately from other forests. However, we should
analyze simultaneously all forests because some vectors of weights may
compensate those vectors which cannot efficiently separate $v_{ij,0}^{(k)}$
and $v_{ij,1}^{(k)}$.

\section{The SDF training and testing}

We apply the greedy algorithm for training the SDF, namely, we train
separately every level starting from the first level such that every next
level uses results of training at the previous level. The training process at
every level consists of two parts. The first part aims to train all trees by
applying all pairs of training examples. This part does not significantly
differ from the training of the original deep forest proposed by Zhou and Feng
\cite{Zhou-Feng-2017}. The difference is that we use pairs of concatenated
vectors $(\mathbf{x}_{i},\mathbf{x}_{j})$ and two classes corresponding to
semantic similarity of the pairs. The second part is to compute the weights
$w^{(t,k)}$, $t=1,...,T_{k}$. This can be done by minimizing the following
objective function over $M$ unit (probability) simplices in $\mathbb{R}%
^{T_{k}}$ denoted as $\Delta_{k}$, i.e., over non-negative vectors
$\mathbf{w}^{(k)}=(w^{(1,k)},...,w^{(T_{k},k)})\in \Delta_{k}$, $k=1,...,M$,
that sum up to one:%
\begin{equation}
\min_{\mathbf{w}}J_{q}(\mathbf{w})=\min_{\mathbf{w}}\sum_{i,j}l(\mathbf{x}%
_{i},\mathbf{x}_{j},y_{ij},\mathbf{w})+\lambda R(\mathbf{w}).
\label{SiamDF_50}%
\end{equation}
Here $\mathbf{w}$ is a vector produced as the concatenation of vectors
$\mathbf{w}^{(k)}$, $k=1,...,M$, $R(\mathbf{w})$ is a regularization term,
$\lambda$ is a hyper-parameter which controls the strength of the
regularization. We define the regularization term as
\[
R(\mathbf{w})=\left \Vert \mathbf{w}\right \Vert ^{2}.
\]
The loss function has to increase values of augmented features $v_{ij,0}%
^{(k)}$ corresponding to the class $c=0$ and to decrease features
$v_{ij,1}^{(k)}$ corresponding to the class $c=1$ for semantically similar
pairs $(\mathbf{x}_{i},\mathbf{x}_{j})$. Moreover, the loss function has to
increase values of augmented features $v_{ij,1}^{(k)}$ corresponding to the
class $c=1$ and to decrease features $v_{ij,0}^{(k)}$ corresponding to the
class $c=0$ for dissimilar pairs $(\mathbf{x}_{i},\mathbf{x}_{j})$.

\subsection{Convex loss function}

Let us denote the set of vectors $\mathbf{w}$ as $\Delta$. In order to
efficiently solve the problem (\ref{SiamDF_50}), the condition of the
convexity of $J_{q}(\mathbf{w})$ in the domain of $\mathbf{w}$ should be
fulfilled. One of the ways for determining the loss function $l$ is to
consider a distance $d(\mathbf{x}_{i},\mathbf{x}_{j})$ between two vectors
$\mathbf{x}_{i}$ and $\mathbf{x}_{j}$ at the $q$-th level. However, we do not
have separate vectors $\mathbf{x}_{i}$ and $\mathbf{x}_{j}$. We have one
vector whose parts correspond to vectors $\mathbf{x}_{i}$ and $\mathbf{x}_{j}%
$. Therefore, this is a distance between elements of the concatenated vector
$\left(  \mathbf{x}_{i}^{(-1)},\mathbf{x}_{j}^{(-1)}\right)  $ obtained at
$q-1$ level and augmented features $\mathbf{v}_{ij}^{(k)}$, $k=1,...,M$, of a
special form. Let us consider the expression for the above distance in detail.
It consists of $M+1$ terms. The first term denoted as $X_{ij}^{(q)}$ is the
Euclidean distance between two parts of the output vector obtained at the
previous level
\[
X_{ij}=\sum_{l=1}^{m}\left(  x_{i,l}^{(-1)}-x_{j,l}^{(-1)}\right)  ^{2}.
\]
Here $x_{i,l}$ is the $l$-th element of $\mathbf{x}_{i}$, $m$ is the length of
the input vector for the $q$-th level or the length of the output vector for
the level with the number $q-1$.

Let us consider elements $v_{ij,0}^{(k)}$ and $v_{ij,1}^{(k)}$ now. We have to
provide the distance between these elements as large as possible taking into
account $y_{ij}$. In particular, if $y_{ij}=0$, then we should decrease the
difference $v_{ij,1}^{(k)}-v_{ij,0}^{(k)}$. If $y_{ij}=1$, then we should
decrease the difference $v_{ij,0}^{(k)}-v_{ij,1}^{(k)}$. Let us introduce the
variable $z_{ij}=-1$ if $y_{ij}=0$, and $z_{ij}=1$ if $y_{ij}=1$. Then the
following expression characterizing the augmented features $v_{ij,0}^{(k)}$
and $v_{ij,1}^{(k)}$ can be written:
\[
\left[  \max \left(  0,z_{ij}\left(  v_{ij,0}^{(k)}-v_{ij,1}^{(k)}\right)
\right)  \right]  ^{2}.
\]
Substituting (\ref{SiamDF_41}) into the above expression, we get next $M$
terms%
\[
\left[  \max \left(  0,~\sum_{t=1}^{T_{k}}P_{ij}^{(t,k)}w^{(t,k)}\right)
\right]  ^{2},\ k=1,...,M,
\]
where%
\[
P_{ij}^{(t,k)}=z_{ij}\left(  p_{ij,0}^{(t,k)}-p_{ij,1}^{(t,k)}\right)  .
\]

Finally, we can write
\begin{equation}
d\left(  \mathbf{x}_{i},\mathbf{x}_{j}\right)  =X_{ij}+\sum_{k=1}^{M}\left[
\max \left(  0,~\sum_{t=1}^{T_{k}}P_{ij}^{(t,k)}w^{(t,k)}\right)  \right]
^{2}. \label{SiamDF_58}%
\end{equation}

So, we have to maximize $d\left(  \mathbf{x}_{i},\mathbf{x}_{j}\right)  $ with
respect to $w^{(t,k)}$ under constraints (\ref{SiamDF_42}). Since $X_{ij}$
does not depend on $w^{(t,k)}$, then we consider the following objective
function%
\begin{equation}
J_{q}(\mathbf{w})=\sum_{i,j}\sum_{k=1}^{M}\left[  \max \left(  0,~\sum
_{t=1}^{T_{k}}P_{ij}^{(t,k)}w^{(t,k)}\right)  \right]  ^{2}+\lambda \left \Vert
\mathbf{w}\right \Vert ^{2}. \label{SiamDF_60}%
\end{equation}

The function $d\left(  \mathbf{x}_{i},\mathbf{x}_{j}\right)  $ is convex in
the interval $[0,1]$ of $w^{(t,k)}$. Then the objective function
$J_{q}(\mathbf{w})$ as the sum of the convex functions is convex too with
respect to weights.

\subsection{Quadratic optimization problem$\allowbreak$}

Let us consider the problem (\ref{SiamDF_60}) under constraints
(\ref{SiamDF_42}) in detail. Introduce a new variable $\xi_{ij}^{(k)}$ defined
as
\[
\xi_{ij}^{(k)}=\max \left(  0,~\sum_{t=1}^{T_{k}}P_{ij}^{(t,k)}w^{(t,k)}%
\right)  .
\]
Then problem (\ref{SiamDF_60}) can be rewritten as%
\begin{equation}
J_{q}(\mathbf{w})=\min_{\xi_{ij}^{(k)},\mathbf{w}}\sum_{i,j}\sum_{k=1}%
^{M}\left(  \xi_{ij}^{(k)}\right)  ^{2}+\lambda \left \Vert \mathbf{w}%
\right \Vert ^{2}, \label{SiamDF_64}%
\end{equation}
subject to (\ref{SiamDF_42})
\begin{equation}
\xi_{ij}^{(k)}\geq \sum_{t=1}^{T_{k}}P_{ij}^{(t,k)}w^{(t,k)},\  \  \xi_{ij}%
^{(k)}\geq0,\  \ (i,j)\in K,\ k=1,...,M. \label{SiamDF_66}%
\end{equation}

We have obtained the standard quadratic optimization problem with linear
constraints and variables $\xi_{ij}^{(k)}$ and $w^{(t,k)}$. It can be solved
by using the well-known standard methods.

It is interesting to note that the optimization problem (\ref{SiamDF_64}%
)-(\ref{SiamDF_66}) can be decomposed into $M$ problems of the form:%
\begin{equation}
J_{q}(\mathbf{w}^{(k)})=\min_{\xi_{ij},\mathbf{w}^{(k)}}\sum_{i,j}\xi_{ij}%
^{2}+\lambda \left \Vert \mathbf{w}^{(k)}\right \Vert ^{2}, \label{SiamDF_70}%
\end{equation}
subject to (\ref{SiamDF_42})
\begin{equation}
\xi_{ij}\geq \sum_{t=1}^{T_{k}}P_{ij}^{(t,k)}w^{(t,k)},\  \  \xi_{ij}%
\geq0,\  \ (i,j)\in K,\ k=1,...,M. \label{SiamDF_72}%
\end{equation}

Indeed, by returning to problem (\ref{SiamDF_64})-(\ref{SiamDF_66}), we can
see that the subset of variables $\xi_{ij}^{(k)}$ and $w^{(t,k)}$ for a
certain $k$ and constraints for these variables do not overlap with the subset
of similar variables for another $k$ and the corresponding constraints. This
implies that (\ref{SiamDF_64}) can be rewritten as
\[
J_{q}(\mathbf{w})=\sum_{k=1}^{M}\min_{\xi_{ij},\mathbf{w}}\sum_{i,j}\left(
\xi_{ij}^{(k)}\right)  ^{2}+\lambda \left \Vert \mathbf{w}\right \Vert ^{2},
\]
and the problem can be decomposed.

So, we solve the problem (\ref{SiamDF_70})-(\ref{SiamDF_72}) for every
$k=1,...,M$ and get $M$ vectors $\mathbf{w}^{(k)}$ which form the vector
$\mathbf{w}$. The above means that the optimal weights are separately
determined for individual forests.

\subsection{A general algorithm for training and the SDF testing}

In sum, we can write a general algorithm for training the SDF (see Algorithm
\ref{alg:SiamDF_4}). Its complexity mainly depends on the number of levels.

Having the trained SDF with computed weights $\mathbf{w}$ for every cascade
level, we can make decision about the semantic similarity of a new pair of
examples $\mathbf{x}_{a}$ and $\mathbf{x}_{b}$. First, the vectors make to be
concatenated. By using the trained decision trees and the weights $\mathbf{w}$
for every level $q$, the pair is augmented at each level. Finally, we get
\[
\mathbf{x}_{ab}^{(Q)}=\left(  \mathbf{x}_{a}^{(Q)},\mathbf{x}_{b}%
^{(Q)}\right)  =\mathbf{v}_{ab}.
\]
Here $\mathbf{v}_{ab}$ is the augmented part of the vector $\mathbf{x}%
_{ab}^{(Q)}$ consisting of elements from subvectors $\mathbf{v}_{0}$ and
$\mathbf{v}_{1}$ corresponding to the class $c=0$ and to the class $c=1$,
respectively. The original examples $\mathbf{x}_{a}$ and $\mathbf{x}_{b}$ are
semantically similar if the sum of all elements from $\mathbf{v}_{0}$ is
larger than the sum of elements from $\mathbf{v}_{1}$, i.e., $\mathbf{v}%
_{0}\cdot \mathbf{1}^{\mathrm{T}}>\mathbf{v}_{1}\cdot \mathbf{1}^{\mathrm{T}}$,
where $\mathbf{1}$ is the unit vector. In contrast to the similar examples,
the condition $\mathbf{v}_{0}\cdot \mathbf{1}^{\mathrm{T}}<\mathbf{v}_{1}%
\cdot \mathbf{1}^{\mathrm{T}}$ means that $\mathbf{x}_{a}$ and $\mathbf{x}_{b}$
are semantically dissimilar and $y_{ab}=1$. We can introduce a threshold
$\tau$ for a more robust decision making. The examples $\mathbf{x}_{a}$ and
$\mathbf{x}_{b}$ are classified as semantically similar and $y_{ab}=0$ if
$\mathbf{v}_{0}\cdot \mathbf{1}^{\mathrm{T}}-\mathbf{v}_{1}\cdot \mathbf{1}%
^{\mathrm{T}}\geq \tau$. The case $0\leq \mathbf{v}_{0}\cdot \mathbf{1}%
^{\mathrm{T}}-\mathbf{v}_{1}\cdot \mathbf{1}^{\mathrm{T}}\leq \tau$ can be
viewed as undeterminable.

\begin{algorithm}
\caption{A general algorithm for training the SDF} \label{alg:SiamDF_4}

\begin{algorithmic}
[1]\REQUIRE Training set $S$ $=\{(\mathbf{x}_{i},\mathbf{x}_{j},y_{ij}%
),\ (i,j)\in K\}$ consisting of $N$ pairs; number of cascade levels $Q$

\ENSURE$\mathbf{w}^{(q)}$, $q=1,...,Q$

\STATE Concatenate $\mathbf{x}_{i}$ and $\mathbf{x}_{j}$ for all pairs of
indices $(i,j)\in K$

\STATE Form the training set $R=\{(\mathbf{x}_{i},\mathbf{x}_{j}%
),y_{ij}),\ (i,j)\in K\}$ consisting of concatenated pairs

\FOR{$q=1$, $q\leq Q$ } \STATE Train all trees at the $q$-th level

\FOR{$k=1$, $k\leq M_q$ } \STATE Compute the weights $\mathbf{w}^{(k)}$ at the
$q$-th level from the $k$-th quadratic optimization problem with the objective
function (\ref{SiamDF_70}) and constraints (\ref{SiamDF_42}) and
(\ref{SiamDF_72})

\ENDFOR

\STATE Concatenate $\mathbf{w}^{(k)}$, $k=1,...,M$, to get $\mathbf{w}$ at the
$q$-th level

\STATE For every $\mathbf{x}_{ij}$, compute $\mathbf{v}_{ij}^{(k)}$ at the
$q$-th level by using (\ref{SiamDF_41}), $k=1,...,M$

\STATE For every $\mathbf{x}_{ij}$, form the concatenated vector
$\mathbf{x}_{ij}$ for the next level by using (\ref{SiamDF_40})

\ENDFOR

\end{algorithmic}
\end{algorithm}

It is important to note that the identical weights, i.e., the gcForest can be
regarded as a special case of the SDF.

\section{Numerical experiments}

We compare the SDF with the gcForest whose inputs are concatenated examples
from series data sets. In other words, we compare the SDF having computed
(trained) weights with the SDF having identical weights. The SDF has the same
cascade structure as the standard gcForest described in \cite{Zhou-Feng-2017}.
Each level (layer) of the cascade structure consists of 2 complete-random tree
forests and 2 random forests. Three-fold cross-validation is used for the
class vector generation. The number of cascade levels is automatically determined.

A software in Python implementing the gcForest is available at
https://github.com/leopiney/deep-forest. We modify this software in order to
implement the procedure for computing optimal weights and weighted averages
$v_{ij,c}^{(k)}$. Moreover, we use pairs of concatenated examples composed of
individual examples as training and testing data.

Every accuracy measure $A$ used in numerical experiments is the proportion of
correctly classified cases on a sample of data. To evaluate the average
accuracy, we perform a cross-validation with $100$ repetitions, where in each
run, we randomly select $N$ training data and $N_{\text{test}}=2N/3$ test data.

First, we compare the SDF with the gcForest by using some public data sets
from UCI Machine Learning Repository \cite{Lichman:2013}: the Yeast data set
(1484 instances, 8 features, 10 classes), the Ecoli data set (336 instances, 8
features, 8 classes), the Parkinsons data set (197 instances, 23 features, 2
classes), the Ionosphere data set (351 instances, 34 features, 2 classes). A
more detailed information about the data sets can be found from, respectively,
the data resources. Different values for the regularization hyper-parameter
$\lambda$ have been tested, choosing those leading to the best results.

In order to investigate how the number of decision trees impact on the
classification accuracy, we study the SDF by different number of trees,
namely, we take $T_{k}=T=100$, $400$, $700$, $1000$. It should be noted that
Zhou and Feng \cite{Zhou-Feng-2017} used $1000$ trees in every forest.

Results of numerical experiments for the Parkinsons data set are shown in
Table \ref{t:SiamDF_4}. It contains the accuracy measures obtained for the
gcForest (denoted as gcF) and the SDF as functions of the number of trees $T$
in every forest and the number $N=100,500,1000,2000$ of pairs in the training
set. It can be seen from Table \ref{t:SiamDF_4} that the accuracy of the SDF
exceeds the same measure of the gcForest in most cases. At that, the
difference is rather large for the small amount of training data. In
particular, the largest differences between accuracy measures of the SDF and
the gcForest are observed by $T=400$, $1000$ and $N=100$. Similar results of
numerical experiments for the Ecoli data set are given in Table
\ref{t:SiamDF_5}. It is interesting to point out that the number of trees in
every forest significantly impacts on the difference between accuracy measures
of the SDF and gcForest. It follows from Table \ref{t:SiamDF_5} that this
difference is smallest by the large number of trees and by the large amount of
training data. If we look at the last row of Table \ref{t:SiamDF_5}, then we
see that the accuracy $0.915$ obtained for the SDF by $T=100$ is reached for
the gcForest by $T=1000$. The largest difference between accuracy measures of
the SDF and the gcForest is observed by $T=100$ and $N=100$. The same can be
seen from Table \ref{t:SiamDF_4}. This implies that the proposed modification
of the gcForest allows us to reduce the training time. Table \ref{t:SiamDF_6}
provides accuracy measures for the Yeast data set. We again can see that the
proposed SDF outperforms the gcForest for most cases. It is interesting to
note from Table \ref{t:SiamDF_6} that the increasing number of trees in every
forest may lead to reduced accuracy measures. If we look at the row of Table
\ref{t:SiamDF_6} corresponding to $N=500$ pairs in the training set, then we
can see that the accuracy measures by $100$ trees exceed the same measures by
larger numbers of trees. Moreover, the largest difference between accuracy
measures of the SDF and the gcForest is observed by $T=1000$ and $N=100$.
Numerical results for the Ionosphere data set are represented in Table
\ref{t:SiamDF_7}. It follows from Table \ref{t:SiamDF_7} that the largest
difference between accuracy measures of the SDF and the gcForest is observed
by $T=1000$ and $N=500$.

The numerical results for all analyzed data sets show that the SDF
significantly outperforms the gcForest by small number of training data
($N=100$ or $500$). This is an important property of the SDF which are
especially efficient when the amount of training data is rather small.%

%TCIMACRO{\TeXButton{B}{\begin{table}[tbp] \centering}}%
%BeginExpansion
\begin{table}[tbp] \centering
%EndExpansion
\caption{Dependence of the accuracy measures on the number of pairs $N$ and on the number of trees $T$ in every forest for the Parkinsons data set }%
\begin{tabular}
[c]{ccccccccc}\hline
$T$ & \multicolumn{2}{c}{$100$} & \multicolumn{2}{c}{$400$} &
\multicolumn{2}{c}{$700$} & \multicolumn{2}{c}{$1000$}\\ \hline
$N$ & gcF & SDF & gcF & SDF & gcF & SDF & gcF & SDF\\ \hline
$100$ & $0.530$ & $0.545$ & $0.440$ & $0.610$ & $0.552$ & $0.575$ & $0.440$ &
$0.550$\\ \hline
$500$ & $0.715$ & $0.733$ & $0.651$ & $0.673$ & $0.685$ & $0.700$ & $0.700$ &
$0.730$\\ \hline
$1000$ & $0.761$ & $0.763$ & $0.778$ & $0.786$ & $0.803$ & $0.804$ & $0.773$ &
$0.790$\\ \hline
$2000$ & $0.880$ & $0.881$ & $0.884$ & $0.895$ & $0.875$ & $0.891$ & $0.887$ &
$0.893$\\ \hline
\end{tabular}
\label{t:SiamDF_4}%
%TCIMACRO{\TeXButton{E}{\end{table}}}%
%BeginExpansion
\end{table}%
%EndExpansion
%

%TCIMACRO{\TeXButton{B}{\begin{table}[tbp] \centering}}%
%BeginExpansion
\begin{table}[tbp] \centering
%EndExpansion
\caption{Dependence of the accuracy measures on the number of pairs $N$ and on the number of trees $T$ in every forest for the Ecoli data set}%
\begin{tabular}
[c]{ccccccccc}\hline
$T$ & \multicolumn{2}{c}{$100$} & \multicolumn{2}{c}{$400$} &
\multicolumn{2}{c}{$700$} & \multicolumn{2}{c}{$1000$}\\ \hline
$N$ & gcF & SDF & gcF & SDF & gcF & SDF & gcF & SDF\\ \hline
$100$ & $0.439$ & $0.530$ & $0.515$ & $0.545$ & $0.590$ & $0.651$ & $0.621$ &
$0.696$\\ \hline
$500$ & $0.838$ & $0.847$ & $0.814$ & $0.823$ & $0.836$ & $0.845$ & $0.821$ &
$0.837$\\ \hline
$1000$ & $0.844$ & $0.853$ & $0.890$ & $0.917$ & $0.888$ & $0.891$ & $0.863$ &
$0.865$\\ \hline
$2000$ & $0.908$ & $0.915$ & $0.895$ & $0.921$ & $0.913$ & $0.915$ & $0.915$ &
$0.915$\\ \hline
\end{tabular}
\label{t:SiamDF_5}%
%TCIMACRO{\TeXButton{E}{\end{table}}}%
%BeginExpansion
\end{table}%
%EndExpansion
%

%TCIMACRO{\TeXButton{B}{\begin{table}[tbp] \centering}}%
%BeginExpansion
\begin{table}[tbp] \centering
%EndExpansion
\caption{Dependence of the accuracy measures on the number of pairs $N$ and on the number of trees $T$ in every forest for the Yeast data set}%
\begin{tabular}
[c]{ccccccccc}\hline
$T$ & \multicolumn{2}{c}{$100$} & \multicolumn{2}{c}{$400$} &
\multicolumn{2}{c}{$700$} & \multicolumn{2}{c}{$1000$}\\ \hline
$N$ & gcF & SDF & gcF & SDF & gcF & SDF & gcF & SDF\\ \hline
$100$ & $0.454$ & $0.500$ & $0.484$ & $0.511$ & $0.469$ & $0.515$ & $0.469$ &
$0.575$\\ \hline
$500$ & $0.682$ & $0.694$ & $0.661$ & $0.673$ & $0.640$ & $0.658$ & $0.622$ &
$0.628$\\ \hline
$1000$ & $0.708$ & $0.711$ & $0.713$ & $0.723$ & $0.684$ & $0.710$ & $0.735$ &
$0.737$\\ \hline
$2000$ & $0.727$ & $0.734$ & $0.714$ & $0.716$ & $0.727$ & $0.739$ & $0.713$ &
$0.722$\\ \hline
\end{tabular}
\label{t:SiamDF_6}%
%TCIMACRO{\TeXButton{E}{\end{table}}}%
%BeginExpansion
\end{table}%
%EndExpansion
%

%TCIMACRO{\TeXButton{B}{\begin{table}[tbp] \centering}}%
%BeginExpansion
\begin{table}[tbp] \centering
%EndExpansion
\caption{Dependence of the accuracy measures on the number of pairs $N$ and on the number of trees $T$ in every forest for the Ionsphere data set}%
\begin{tabular}
[c]{ccccccccc}\hline
$T$ & \multicolumn{2}{c}{$100$} & \multicolumn{2}{c}{$400$} &
\multicolumn{2}{c}{$700$} & \multicolumn{2}{c}{$1000$}\\ \hline
$N$ & gcF & SDF & gcF & SDF & gcF & SDF & gcF & SDF\\ \hline
$100$ & $0.515$ & $0.515$ & $0.535$ & $0.555$ & $0.530$ & $0.555$ & $0.535$ &
$0.540$\\ \hline
$500$ & $0.718$ & $0.720$ & $0.723$ & $0.758$ & $0.713$ & $0.740$ & $0.715$ &
$0.760$\\ \hline
$1000$ & $0.820$ & $0.830$ & $0.837$ & $0.840$ & $0.840$ & $0.860$ & $0.885$ &
$0.895$\\ \hline
$2000$ & $0.915$ & $0.920$ & $0.905$ & $0.905$ & $0.895$ & $0.895$ & $0.910$ &
$0.910$\\ \hline
\end{tabular}
\label{t:SiamDF_7}%
%TCIMACRO{\TeXButton{E}{\end{table}}}%
%BeginExpansion
\end{table}%
%EndExpansion

It should be noted that the multi-grained scanning proposed in
\cite{Zhou-Feng-2017} was not applied to investigating the above data sets
having relatively small numbers of features. The above numerical results have
been obtained by using only the forest cascade structure.

\begin{figure}
[ptb]
\begin{center}
\includegraphics[
%natheight=4.310400in,
%natwidth=3.874400in,
height=2.1008in,
width=1.8893in
]%
{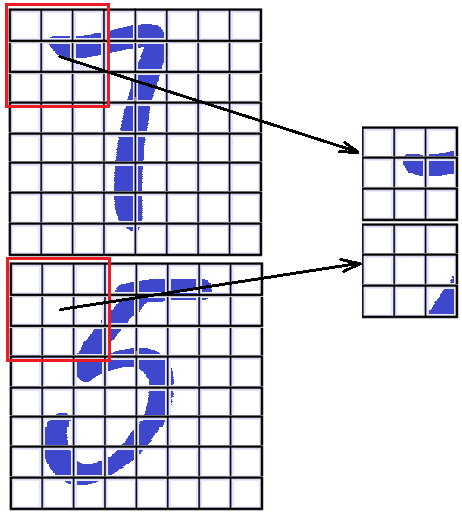}%
\caption{The multi-grained scanning scheme for concatenated examples}%
\label{fig:scan_win}%
\end{center}
\end{figure}

When we deal with the large-scale data, the multi-grained scanning scheme
should be use. In particular, for analyzing the well-known MNIST data set, we
used the same scheme for window sizes as proposed in \cite{Zhou-Feng-2017},
where feature windows with sizes $\left \lfloor d/16\right \rfloor $,
$\left \lfloor d/9\right \rfloor $, $\left \lfloor d/4\right \rfloor $ are chosen
for $d$ raw features. We study the SDF by applying the MNIST database which is
a commonly used large database of $28\times28$ pixel handwritten digit images
\cite{LeCun-etal-1998}. It has a training set of 60,000 examples, and a test
set of 10,000 examples. The digits are size-normalized and centered in a
fixed-size image. The data set is available at
http://yann.lecun.com/exdb/mnist/. The main problem in using the multi-grained
scanning scheme is that pairs of the original examples are concatenated. As a
result, the direct scanning leads to scanning windows covering some parts from
every example belonging to a concatenated pair, which do not correspond the
images themselves. Therefore, we apply the following modification of the
multi-grained scanning scheme. Two identical windows simultaneously scan two
concatenated images such that pairs of feature windows are produced due to
this procedure, which are concatenated for processing by means of the forest
cascade. Fig. \ref{fig:scan_win} illustrates the used procedure. Results of
numerical experiments for the MNIST data set are shown in Table
\ref{t:SiamDF_8}. It can be seen from Table \ref{t:SiamDF_8} that the largest
difference between accuracy measures of the SDF and the gcForest is observed
by $T=1000$ and $N=100$. It is interesting to note that the SDF as well as the
gcForest provide good results even by the small amount of training data. At
that, the SDF outperforms the gcForest in the most cases.%

%TCIMACRO{\TeXButton{B}{\begin{table}[tbp] \centering}}%
%BeginExpansion
\begin{table}[tbp] \centering
%EndExpansion
\caption{Dependence of the accuracy measures on the number of pairs $N$ and on the number of trees $T$ in every forest for the MNIST data set}%
\begin{tabular}
[c]{ccccccccc}\hline
$T$ & \multicolumn{2}{c}{$100$} & \multicolumn{2}{c}{$400$} &
\multicolumn{2}{c}{$700$} & \multicolumn{2}{c}{$1000$}\\ \hline
$N$ & gcF & SDF & gcF & SDF & gcF & SDF & gcF & SDF\\ \hline
$100$ & $0.470$ & $0.490$ & $0.520$ & $0.520$ & $0.570$ & $0.585$ & $0.530$ &
$0.570$\\ \hline
$500$ & $0.725$ & $0.735$ & $0.715$ & $0.715$ & $0.695$ & $0.700$ & $0.670$ &
$0.670$\\ \hline
$1000$ & $0.757$ & $0.770$ & $0.755$ & $0.760$ & $0.775$ & $0.780$ & $0.830$ &
$0.840$\\ \hline
\end{tabular}
\label{t:SiamDF_8}%
%TCIMACRO{\TeXButton{E}{\end{table}}}%
%BeginExpansion
\end{table}%
%EndExpansion
%

%TCIMACRO{\FRAME{ftbpFU}{1.8893in}{2.1008in}{0pt}{\Qcb{The multi-grained
%scanning scheme for concatenated examples}}{\Qlb{fig:scan_win}}%
%{weighted_class_vector_gen_4.png}{\special{ language "Scientific Word";
%type "GRAPHIC";  maintain-aspect-ratio TRUE;  display "USEDEF";
%valid_file "F";  width 1.8893in;  height 2.1008in;  depth 0pt;
%original-width 3.8744in;  original-height 4.3104in;  cropleft "0";
%filename 'Weighted_Class_vector_gen_4.png';file-properties "XNPEU";}}}%
%BeginExpansion
%EndExpansion

An interesting observation has been made during numerical experiments. We have
discovered that the variable $z_{ij}$, initially taking the values $-1$ for
$y_{ij}=0$ and $1$ for $y_{ij}=1$, can be viewed as a tuning parameter in
order to control the number of the cascade levels used in the training process
and to improve the classification performance of the SDF. One of the great
advantages of the gcForest is its automatic determination of the number of
cascade levels. It is shown by Zhou and Feng \cite{Zhou-Feng-2017}, that the
performance of the whole cascade is estimated on validation set after training
a current level. The training procedure in the gcForest terminates if there is
no significant performance gain. It turns out that the value of $z_{ij}$
significantly impact on the number of cascade levels if to apply the
termination procedure implemented in the gcForest. Moreover, we can adaptively
change the values of $z_{ij}$ with every level. It has been revealed that one
of the best change of $z_{ij}$ is $z_{ij}^{(q)}=2z_{ij}^{(q-1)}$, where
$z_{ij}^{(1)}=-1$ for $y_{ij}=0$ and $1$ for $y_{ij}=1$. Of course, this is an
empirical observation. However, it can be taken as a direction for further
improving the SDF.

\section{Conclusion}

One of the implementations of the SDF has been represented in the paper. It
should be noted that other modifications of the SDF can be obtained. First of
all, we can improve the optimization algorithm by applying a more complex loss
function and computing optimal weights, for example, by means of the
Frank-Wolfe algorithm \cite{Frank-Wolfe-1956}. We can use a more powerful
optimization algorithm, for example, an algorithm proposed by Hazan and Luo
\cite{Hazan-Luo-2016}. Moreover, we do not need to search for the convex loss
function because there are efficient optimization algorithms, for example, a
non-convex modification of the Frank-Wolfe algorithm proposed by Reddi et al.
\cite{Reddi-etal-2016}, which allows us to solve the considered optimization
problems. The trees and forests can be also replaced with other classification
approaches, for example, with SVMs and boosting algorithms. However, the above
modifications can be viewed as directions for further research.

The linear combinations of weights for every forest have been used in the SDF.
However, this class of combinations can be extended by considering non-linear
functions of weights. Moreover, it turns out that the weights of trees can
model various machine learning peculiarities and allow us to solve many
machine learning tasks by means of the gsForest. This is also a direction for
further research.

It should be noted that the weights have been restricted by constraints of the
form (\ref{SiamDF_42}), i.e., the weights of every forest belong to the unit
simplex whose dimensionality is defined by the number of trees in the forest.
However, numerical experiments have illustrated that it is useful to reduce
the set of weights in some cases. Moreover, this reduction can be carried out
adaptively by taking into account the classification error at every level. One
of the ways for adaptive reduction of the unit simplex is to apply imprecise
statistical models, for example, the linear-vacuous mixture or imprecise
$\varepsilon$-contaminated models proposed by Walley \cite{Walley91}. This
study is also a direction for further research.

We have considered a weakly supervised learning algorithm when there are no
information about the class labels of individual training examples, but we
know only semantic similarity of pairs of training data. It is also
interesting to extend the proposed ideas on the case of fully supervised
algorithms when only the class labels of individual training examples are
known. The main goal of fully supervised distance metric learning is to use
discriminative information in distance metric learning to keep all the data
samples in the same class close and those from different classes separated
\cite{Mu-Ding-2013}. Therefore, another direction for further research is to
adapt the proposed algorithm for the case of available class labels.

\section*{Acknowledgement}

The reported study was partially supported by RFBR, research project No. 17-01-00118.

%%%\bibliographystyle{plain}
%%%%\bibliography{Autoencoder,Classif_bib,Cluster_bib,Deep_Forest,Interv_NN,IntervalClass,MYUSE,Robots}

\end{document}